\documentclass{llncs}
\usepackage{tabularx,graphicx,subfigure}
\usepackage{xcolor}
\newcommand{\vs}{\vspace{-3pt}} 
\newcommand{\citebvs}[1]{\cite{#1}}

\begin{document}

\title{Generating summaries tailored to target characteristics}

\author{
Kushal Chawla$^\ddag$, Hrituraj Singh$^\dag$, Arijit Pramanik$^*$, Mithlesh Kumar$^\#$, \\Balaji Vasan Srinivasan$^\ddag$}
\institute{$^\ddag$Adobe Research, Bangalore, India\\
		 $^\dag$Indian Institute of Technology, Roorkee\\
		 $^*$Indian Institute of Technology, Bombay\\
		 $^\#$Indian Institute of Technology, Kanpur\\
           kchawla@adobe.com,balsrini@adobe.com}

\maketitle

\begin{abstract}
Recently, research efforts have gained pace to cater to varied user preferences while generating text summaries. While there have been attempts to incorporate a few handpicked characteristics such as length or entities, a holistic view around these preferences is missing and crucial insights on why certain characteristics should be incorporated in a specific manner are absent. With this objective, we provide a categorization around these characteristics relevant to the task of text summarization: one, focusing on what content needs to be generated and second, focusing on the stylistic aspects of the output summaries. We use our insights to provide guidelines on appropriate methods to incorporate various classes characteristics in sequence-to-sequence summarization framework. Our experiments with incorporating topics, readability and simplicity indicate the viability of the proposed prescriptions.
\end{abstract}

\section{Introduction}\vs\vs
Automatic text summarization~\cite{nenkova2011automatic} is the task of generating a summary of an input document while retaining the key aspects. Such a summary helps in presenting the important content from a long input text in a succinct form for quick information consumption. Traditional methods for summarization \cite{nenkova2011automatic} extract key sentences from the source text to construct an `extractive' summary. Recent efforts towards `abstractive' summarization have geared towards generating human-like, paraphrased summaries from the input article~\cite{nallapati2017summarunner,see2017get}.

While these algorithms allow for the generation of a single summary, it is often desirable to generate summary variants tailored towards specific characteristics. For instance, readers may be interested in summaries of different lengths or might want to focus on specific entities/topics from the input text~\cite{fan2017controllable,krishna2018generating,wang2018reinforced}. Depending on different age groups of the readers, they might prefer formal/informal variants of the summary~\cite{niu2017study}. Irrespective of the application context, it has been shown that incorporating these characteristics at the time of generation can yield more contextual summaries \cite{krishna2018vocabulary}.

Recent works have proposed different ways to incorporate target characteristics at the time of summary generation: introducing modifications to the dataset~\cite{fan2017controllable,krishna2018generating}, architecture~\cite{niu2017study}, learning objectives~\cite{paulus2018deep} or the decoder probabilities \cite{krishna2018vocabulary}. However, all these attempts handpick a few characteristics and propose ways to incorporate them. In the absence of appropriate insights, it is unclear as to why these methodologies work in tuning the summaries towards the chosen characteristics and why the same cannot be extended for other characteristics. In this work, our objective is to gain a holistic understanding around these additional constraints, centering on the task of text summarization.

Taking a step in this direction, we propose a categorization of these characteristics into 1) \textbf{content-specific}, which primarily focus on what content is presented pivoting on the semantics or information presented in the output summary and 2) \textbf{style-specific} focusing on the stylistic expressions pivoting on the linguistic presentation in the output summary. Through a comparative evaluation of various existing and proposed methods, we further prescribe guidelines to help choosing the right framework for tailoring these categories of characteristics.
%The notion of style and the associated nomenclature is quite convoluted and has been confused often in the literature~\cite{tikhonov2018wrong}. Hence, we explain the key terms we use in this paper in Section \ref{sec:cs}. This allows us to scope our problem setting and describe our categorization of content and style. We have further utilized this to classify prior research efforts in Section \ref{relatedwork}. As an attempt to incorporate these characteristics into the summarization process, we propose a general methodology and provide guidelines for both categories. To show the validity of our propositions, we evaluate on a representative content-specific characteristics-Topics and two style-specific characteristics, namely Readability and Simplicity (Sections \ref{sec:content} and \ref{sec:style}).
Our primary contribution is providing a categorization of target characteristics as content and style specific, towards a holistic understanding of tailored summary generation. %Content-specific characteristics mainly govern the semantics or information presented in the output and style-specific characteristics are related to various aspects of style. Summarization itself can be considered as a content-specific characteristic since it extracts key information from the input content to be expressed in the summary. Additionally, there may be other constraints which must be tailored in the summary, e.g. central \textit{topic} in the summary. We also explore tailoring style-specific characteristics such as \textit{readability} and \textit{simplicity} which tailor the linguistic aspects in the generated summary. We provide guidelines for characteristic tailoring through a comparison across multiple prior techniques. 
Additionally, we propose an attention-boosting approach to improve tailoring of content-specific characteristics and a policy gradient based algorithm to incorporate stylistic characteristics in summaries.

\vs\vs\section{Related Work}\vs\vs 
\label{relatedwork}
Traditional methods for summarization~\cite{nenkova2011automatic} extract key sentences from the source text to construct an `extractive' summary. Features like descriptiveness of words and word frequencies have been explored to choose the summary sentences. However, humans summarize by understanding the content and paraphrasing the understood content into a summary. Extractive summarization is hence unable to produce `human-like' summaries. This has led to efforts towards `abstractive' summarization, which paraphrase summaries from input article content.

Early attempts at abstractive summarization created summary sentences either based on templates~\cite{wang2013domain,genest2011framework} or used ILP-based sentence compression techniques \cite{filippova2010multi,berg2011jointly,banerjee2015multi}. With the advent of deep sequence to sequence models~\cite{sutskever2014sequence}, attention-based neural models have been proposed for summarizing long sentences~\cite{rush2015neural,chopra2016abstractive}. These approaches were further improved by incorporating Abstract Meaning Representations~\cite{takase2016neural} and using hierarchical encoding networks~\cite{nallapati2016abstractive}. %All these efforts involved two sentence-level datasets, Gigaword and DUC-2004.
More recent approaches~\cite{nallapati2017summarunner,see2017get} have focused on large scale datasets for summarization such as CNN/DailyMail corpus~\cite{hermann2015teaching,nallapati2016abstractive}. Gulcehre et al. ~\citebvs{gulcehre2016pointing} introduced the ability to copy out-of-vocabulary words from the article to incorporate rarely seen words like names in the generated text. Tu et al. \citebvs{tu2016modeling} included the concept of coverage, to prevent the models from repeating the same phrases while generating a sentence. See et al. \citebvs{see2017get} proposed a pointer-generator framework which incorporates these improvements, and also learns to switch between generating new words and copying words from the source article. %We build our work on the pointer-generator framework.

%\subsection{Incorporating additional constraints in summary generation}
Although research has primarily focused on unconditioned abstractive text summarization, there have been some recent efforts to \textbf{incorporate a variety of additional constraints} into the generation algorithm. Fan et al. ~\citebvs{fan2017controllable} use explicit indicators to control the length of the output summary, imposing a constraint on how detailed the output needs to be. They extend the same technique to also control the entities which must be focused on while generating the summary. Krishna et al. ~\citebvs{krishna2018generating} generated topic-oriented summaries by using an indicator topic-vector along with the input representation. Each of these approaches aim to control the \textit{information} presented in the generated summary and therefore modifies the attention distribution to focus on appropriate parts of the text as dictated by the target characteristics. We group these approaches as \textbf{content-specific }characteristic tuning.

Another direction of efforts have attempted to incorporate aspects like sentiments or tense using generative models like variational auto-encoders~\cite{hu2017toward} or using adversarial training~\cite{shen2017style}. Focusing on politeness, Sennrich et al. ~\citebvs{sennrich2016controlling} propose modifications to neural machine translation setup to generate polite variants. Ficler et al. ~\citebvs{ficler2017controlling} propose the use of a conditional language model to generate text with variations such as descriptiveness, personal and sentiment simultaneously. More recently, generating text with varying levels of formality was studied in Machine Translation~\cite{niu2017study,niu2018multi}. Oraby et al. ~\citebvs{oraby2018controlling} attempt to control personality dimensions in generation, namely, agreeable, disagreeable, conscientious, unconscientious and extravert by using indicator tokens and stylistic encodings. Krishna et al. ~\citebvs{krishna2018vocabulary} modify the decoding algorithm to produce readable and simple summaries. Each of these exploration focus primarily of tuning the linguistic presentation of the generated text and we group these as \textbf{style-specific} characteristic tuning.

Policy learning based approaches~\cite{paulus2018deep} have shown promise to control several qualitative characteristics explicitly and can potentially be used for both content and style specific characteristics. While it has been successfully deployed for metrics like ROUGE \cite{paulus2018deep}, we show its applicability to style-specific characteristics by proposing a policy gradient framework for readability and simplicity.

\noindent\textbf{Content vs Style:} The notion of style and the associated nomenclature is quite convoluted in the literature~\cite{tikhonov2018wrong}. Approaches in style transfer try to obtain independent latent representations for style and semantics~\cite{artetxe2017unsupervised},\cite{han2017unsupervised}, \cite{shen2017style}, \cite{xu2018unpaired}, \cite{prabhumoye2018style}, \cite{zhang2018shaped} of the content. Using this interpretation, we see the output of a text generation system as a combination of the semantics of the information which is being presented and the style associated with that content. However, unlike these approaches that learn the notion of style implicitly from the available corpora, we fragment style across a set of dimensions such as readability, simplicity, etc. These are referred to as various aspects of style aligning with ad-hoc approaches towards style transfer as described in \cite{tikhonov2018wrong}. 

%These characteristics can be specific to both the content creators and the target audience. For instance, depending on how detailed output the audience is interested in, one may want to control the length of the output summary. On the other hand, characteristics like `extrovertism' refer to controlling the `extrovert' personality of the creator \cite{oraby2018controlling} expressed in the content. In this work, we scope our categorization to only the characteristics significant from the point of view of the target audience.

%We are now in a position to categorize various preferences of the target audience for the task of abstractive text summarization. 

\vs\vs\section{Pointer-Generator Framework}\vs\vs
We base all our explorations on the pointer generator network \cite{see2017get}. However, all our findings and insights are generic and can be extended to any other frameworks without loss of generality. We describe the pointer generator framework here for the sake of completion and please refer to \cite{see2017get} for more details on the framework. The pointer generator network \cite{see2017get} consists of an encoder and a decoder, both based on LSTM architecture. Given an input article, the encoder takes the embedding vectors of each word in the source text $A$ and computes the encoder hidden states $h_1,h_2,..h_n$. The final hidden state is passed to a decoder, which computes a hidden state $s_t$ at each decoding step and calculates an attention distribution $a^t$ over all words as $a^t = \rm{softmax}(e^t)$, where,
\begin{equation}
e_i^t = v^T \tanh(W_h h_i + W_s s_t + b_{att})
\label{eq:attention}
\end{equation}
where $v, W_h, W_s$ and $b_{att}$ are model parameters to be trained. The attention is a probability distribution over words in the source text, which aids the decoder in generating the next word in the summary using words in the source text with higher attention. The context vector $h_t^* = \sum\limits_{i=1}^n a_i^t h_i,$ is a weighted sum of the encoder hidden states (weighted on $a^t_i$ the attention on the $i^{th}$ input word at $t^{th}$ step) and is used to determine the next word to be generated. The attention distribution allows the network to focus on specific parts of the input as the output summary is generated. To tailor a summary to various content-specific characteristics, it is important to modify this attention distribution to focus on the appropriate parts of the input text as required by the characteristics tuned. 

At each decoding step, the decoder also gets the last word $y_t$ in the summary generated so far and computes a scalar $p_{gen}$ denoting the probability of generating a new word from the vocabulary, $p_{gen} = \sigma (w_h^Th_t^*+w_s^Ts_t+w_y^Ty_t+b_{gen})$, where $w_h, w_s, w_y, b_{gen}$ are trained vectors. The network probabilistically decides based on $p_{gen}$, whether to generate a new word from the vocabulary or copy a word from the source text using the attention distribution. For each word $w$ in the vocabulary, the model calculates $P_{vocab}(w)$, the probability of the word getting generated next. For each word $w'$ in the input article, its total attention received yields its probability of being copied. Since  some words occur in the vocabulary and also the input article, they will have non-zero probabilities of being newly generated as well as being copied. Hence, the total probability of $w$ being the next word generated in the summary(denoted by $\mathbf{p}$) is given by,
\begin{equation}
\mathbf{p}(w) = p_{gen}P_{vocab}(w) + (1-p_{gen})\sum_{i:w_i = w}a_i^t\label{eq:pw}
\end{equation}
The second term allows the framework to choose a word to copy from the input text using the attention distribution. The pointer-generator network further employs a coverage mechanism to encourage diversity in attention distributions over time steps. The training loss is set to be the average negative log-likelihood of the ground truth summaries. The model is trained using back-propagation and the Adagrad gradient descent algorithm \cite{duchi2011adaptive}. Since the stylistic characteristics deal with specific expressions of the output text, it can be tailored by modifying $p(w)$ to incorporate the corresponding stylistic preferences. For more complex characteristics, as we show, it is possible to define a reinforcement learning based loss appended to the training loss to tailor the specific characteristics.

\vs\vs
\section{Content-specific Characteristics}
\label{sec:content}\vs\vs
Content-specific characteristics primarily govern what content needs to be presented in the output summary. For the rest of the paper, we illustrate the needs and modeling for content based characteristics with topical tailoring. However, the proposed approach can be extended to other content characteristics like entity-centric tailoring, etc.  

Often, the whole content of the article may not be relevant to the readers and may prefer specific elements of the input to be summarized. For instance, a sports enthusiast may only be interested in content concerning that domain, or a surgeon may only be interested in health-related content. This calls for a need to generate multiple summary variants taking this information into account. Table \ref{tab:topic_examples} shows a particular instance from our dataset which talks about both Politics and Military. If a reader is interested only in Politics, the baseline summary generated by pointer-generator (\textbf{PGen}) model does not refer to politics and hence fails to meet the needs. %\todo{edit the table here.}

\begin{table}[bth]
	\begin{center}\small\vs\vs
		\begin{tabularx}{\linewidth}{|X|}
			\hline
			\textbf{Article:}  bernie sanders , my vermont senator and , indeed , a friend of many years , is now running for president . he noted at his announcement -lrb- with a familiar note of wise irony -rrb- : `` people should not underestimate me . '' to most americans , of course , sen. bernie sanders is only a name , if that . he is barely known to the general public , which makes him a very long shot indeed to win election to the highest office in the nation . -lrb- cnn -rrb- he was impressively polite and bright in the eyes of his boyhood teachers , an encourager of his college friends . he was a docile captured killer in the care of paramedics tending to his gunshot wounds . dzhokhar `` jahar '' tsarnaev 's defense team is seeking to spare him from a death sentence for his part two years ago in the boston marathon bombings and murder of an mit police officer ...\\ \hline
			\textbf{PGen:} dzhokhar `` jahar '' tsarnaev 's defense team is seeking to spare him from a death sentence . he was a docile captured killer in the care of paramedics tending to his gunshot wounds . tsarnaev was convicted april 8 on all 30 counts , including 17 that carry a possible death penalty .\\ \hline
			\textbf{Token-based (mixed) \cite{krishna2018generating} + Attention Boosting; Politics:} sen. bernie sanders is running for president . he is barely known to the general public , which makes him a very long shot indeed . he is barely known to the general public , which makes him a very long shot indeed .\\
			\textbf{Token-based (mixed) \cite{krishna2018generating} + Attention Boosting; Military:} dzhokhar `` jahar '' tsarnaev 's defense team is seeking to spare him from a death sentence . he was convicted april 8 on all 30 counts , including 17 that carry a possible death penalty . a paramedic testified wednesday that it was common for patients in shock to become agitated.\\\hline
		\end{tabularx}
	\end{center}\vs\vs
	\caption{Sample output topic-tailored summaries generated by Token-Based approach trained on CNN-DM-mixed dataset. We show just the top few sentences in the input article, in the interest of space.\vs\vs\vs\vs\vs\vs}
	\label{tab:topic_examples}
\end{table}

Sequence-to-sequence learning models have been shown to understand where to look in the input through attention mechanisms which is then used for output generation. Tailoring content-specific characteristics would require this attention to be tuned to focus on the relevant parts of the input (e.g. the relevant parts of the input talking about a topic of interest) to generate the desired output. This requires the model to be taught (either explicitly or implicitly) where to attend in the input article to tailor the summary appropriately. 

One possibility is to maintain explicit indicators for each category of the characteristic (e.g. for each topic), allowing the model to learn where to pay more attention directly from the data. Fan et al. ~\cite{fan2017controllable} propose to use such indicator tokens to tune characteristics such as length or desired entities. When training on an (article, summary) pair belonging to a particular bin, a token indicating the characteristic (topic in this case) represented by the summary is added to the beginning of the input article. While decoding an unseen article, the framework can generate multiple summary variants based on what token is prepended to the input word sequence. Internally, the model uses the token to learn a conditioned space of parameters, ensuring appropriate attention tuning to generate the summary with corresponding tailoring. We refer to this approach as \textbf{Token-Based} in our experiments.

A key requirement to make the model learn these intricacies is that the training data should contain sufficient samples under each category. However, there could be a skew in the dataset, which calls for alternate approach to tackle these characteristics. To deal with this problem, Krishna et al. ~\cite{krishna2018generating} create a separate dataset where the model sees multiple summary variants for the same input article by mixing multi-topic articles. Given such a dataset, the vocabulary tokens can be used to guide the learning process towards a topic specific attention with a skewed-dataset. This method is called as \textbf{Token-Based (mixed)} which is trained on such an interspersed dataset.

While \cite{fan2017controllable} and \cite{krishna2018generating} use token based approaches implicitly teach the networks by taking advantage of the diversity in the training data, we propose an alternative to ``explicitly'' boost the attention distributions (referred to as \textbf{Attn-Boost}), restricting the model to focus on some parts of the input more than others. More formally, we modify Eq. \ref{eq:attention} from the pointer generator as,
\begin{equation}
e_i^{t'} = \beta_{i}v^T \tanh(W_h h_i + W_s s_t + b_{att}),
\label{eq:attention2}
\end{equation}
where $v, W_h, W_s$ and $b_{att}$ are trainable model parameters as before and use this $e_i$ to compute the attention $a^t$. $\beta_{i}$ is a word specific attention boosting parameter. This explicitly teaches the model to pay more attention towards specific words than others. We leverage topic specific word lists curated by \citebvs{krishna2018generating} and select the top $k$ ($=5$ in our experiments) sentences from the input article, which are most related to the target topic. We explicitly boost $\beta_{i}$s of all the words in these sentences using the topic confidence measures as used by \citebvs{krishna2018generating}.

To draw more insights on each of these approaches, we evaluate it on the task of topic-based summarization on the CNN/DailyMail (\textbf{CNN-DM}) dataset \cite{hermann2015teaching,nallapati2016abstractive}. The dataset consists of $287,226$ training, $13,368$ validation and $11,490$ test instances. The articles have an average length of $781$ tokens and multi-sentence summaries with average length of $56$ tokens. We use the vanilla pointer generator (\textbf{PGen}) as the baseline for all our experiments retaining the hyper-parameters by See et al. ~\cite{see2017get}. We train the \textbf{Token-Based} model \cite{fan2017controllable} for topics by categorizing the ground-truth summaries into $9$ topics: business, education, entertainment, health, military, politics, social, sports and technology (extending the setup by \citebvs{krishna2018generating}) and prepending the topic to the input article while training. 

Following~\citebvs{krishna2018generating}, we also have a setup where we intersperse articles from different topics in \textbf{CNN-DM}, resulting in multiple topic-specific ground truth summaries for the same article. Using the $9$ topics as before, the model now sees multiple summaries for the same article. %We call this dataset as \textbf{CNN-DM-mixed}.
There are $103,666$ article--summary-topic tuples for training, $4,720$ tuples for validation and $3,964$ tuples in the test dataset. %Finally, we also evaluate a combination of Token-Based approaches with the proposed explicit attention boosting method described above. 

To evaluate the generation quality of all the approaches, we compare them on ROUGE $1$, $2$ and $L$ F1-score. %To assess whether we are able to generate topic-specific summaries, we define two metrics. 
Note that generating summaries for all the topics may not make sense for the same input article, particularly when the article does not talk about the target topic. Hence, for each article in the test set, we generate the summary corresponding to the target topic defined by the ground truth summary. Then, we use the topic specific word lists ~\cite{krishna2018generating} to get the top-$1$ and top-$3$ topics in the decoded summaries. The fraction of times the target topic lies in top-$1$ and top-$3$ topics of the decoded summaries, defines the \textbf{Top1} and \textbf{Top3} accuracies of the various setups.

\begin{table}[th]
	\centering\vs\vs\vs
	\scriptsize
	\begin{tabularx}{\linewidth}{|X||c|c|c||c|c|}
		\hline
		\textbf{Method} & \multicolumn{3}{c||}{\textbf{ROUGE F-score}} & \multicolumn{2}{c|}{\textbf{\% accuracy}}  \\ \hline
		& \textbf{1} & \textbf{2} & \textbf{L} & \textbf{Top-1} & \textbf{Top-3} \\ \hline
		\textbf{PGen \cite{see2017get}} &$23.13$ &$7.59$ & $21.01$&$35.00$ & $56.39$\\ \hline
		\textbf{Token-Based \cite{fan2017controllable}} & $26.53$ & $9.05$ & $23.98$ & $37.52$ &  $63.36$ \\ \hline
		\textbf{Attn-Boost (Proposed)} &$22.05$ &$7.30$ & $19.95$&$40.06$ & $62.85$\\ \hline
		\textbf{Token-Based \cite{fan2017controllable} + Attn-Boost} & $25.94$ & $8.63$ & $23.35$ & $40.26$ & $64.77$  \\ \hline
		%\textbf{Token-Based (mixed)} &$29.82$ & $11.53$& $26.98$ & $59.26$ & $78.33$ \\ \hline
		\textbf{Token-Based (mixed) \cite{krishna2018generating} + Attn-Boost} & $\mathbf{28.30}$ & $\mathbf{10.46}$ & $\mathbf{25.45}$ & $\mathbf{55.86}$ & $\mathbf{76.19}$ \\ \hline
	\end{tabularx}
	\caption{\label{tab:topic_results} Performance of proposed methodologies for generating topic-tuned summaries on the \textbf{CNN-DM-mixed} test dataset.\vs\vs\vs\vs\vs}
\end{table}

Table \ref{tab:topic_results} summarizes the results of our methods for generating topic-oriented summaries. We observe that boosting attention values explicitly shows improvement in topic percentage accuracies but it suffers a decline in quality based on Rouge scores. This is expected since the explicit topic attention would make the model attend to parts of documents that are different from the ground truth summary. 
On the other hand, token based approach improves on ROUGE with a lesser topical accuracy. A combined framework of token-based and attention boosting yields the best performance across both ROUGE and topical accuracy metrics. This is perhaps because in the combined setup, the model learns more intricacies implicit in the data along with explicit attention to the topics - thus getting the best of both frameworks. 
When we train the same setup with the mixed dataset by \citebvs{krishna2018generating}, both ROUGE and the topical accuracies improve, suggesting the importance of the diversity in data for the network to implicitly learn attention patterns. %Giving the target topic along with the input article, allows the model to generate content in accordance with it, outputting summaries more aligned with the ground truth and hence, improving on the ROUGE metric. Topic information also improves on Top-1 and Top-3 accuracies. Although the improvements with using \textbf{Token-Based} on \textbf{CNN-DM} are only marginal, using \textbf{CNN-DM-mixed}, shows a much higher gain in performance. Being trained on a dataset with different output summaries for the same input, the latter is able to focus only on specific aspects of the article. Augmenting the Token-Based approach with the explicit boosting, we find improvements for the method trained on \textbf{CNN-DM}, but not on the one trained on \textbf{CNN-DM-mixed} dataset.
%\todo{add combined results to the table and discuss here.}

\begin{figure*}[bth]
	\centering
	\subfigure[Politics]{
		\includegraphics[width=0.9\linewidth]{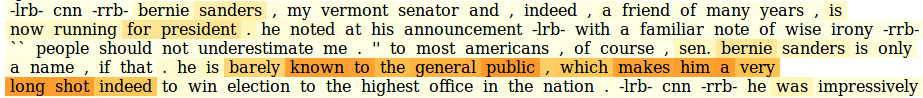}}
	\subfigure[Military]{
		\includegraphics[width=0.9\linewidth]{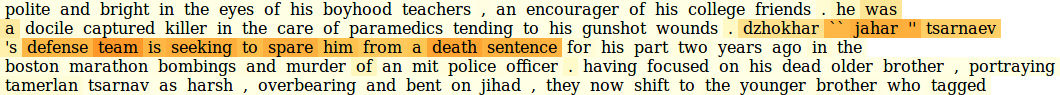}}
	\caption{Attention Distribution Over the Source Article for different Target Topics.\vs\vs\vs\vs}
	\label{fig:attention_dists}
\end{figure*}

We show generated summaries from the token based approach in Table \ref{tab:topic_examples} for a particular instance from the testing dataset. The article was created by combining two articles from Politics and Military domain. The proposed approach appended with token-based framework on \textbf{CNN-DM-mixed} dataset is able to generate topic-specific variants, while the \textbf{PGen} approach fails to meet the requirement towards both the topics. Figure \ref{fig:attention_dists} shows the average attention on the most attended parts of the input for the same instance, by the token based model trained on \textbf{CNN-DM-mixed}. When generating a 'Politics' oriented summary, the attention is on words like 'president', 'bernie sanders' and 'general public' showing the bias towards political phrases. On the other hand, when the target topic is 'Military', the focus/attention shifts to 'death sentence' and 'defence team'. 

Our evaluations show that tuning content based characteristics can be achieved by modifying the attention of the network - either implicitly or explicitly. Implicit attention modification uses a token based approach \cite{fan2017controllable}, but relies on the diversity of the characteristics in the data. Where not available, feeding off of an interspersed dataset to artificially infuse diversity \cite{krishna2018generating} is beneficial. By combining an interspersed dataset with explicitly attention boosting framework, the model is able to tune the characteristics better by learning where to attend and where-not to attend while tuning the content-based characteristics.

\vs\vs\section{Stylistic Characteristics}
\label{sec:style}\vs\vs
Next, we focus on incorporating stylistic characteristics in the generated summary. Style-specific preferences ensure that the content is served appropriately to the target audience. In this direction, prior work has focused on incorporating dimensions such as sentiment~\cite{hu2017toward}, descriptiveness~\cite{ficler2017controlling}, formality~\cite{niu2017study} and many more across various tasks in text generation. We describe our methodology below to incorporate such characteristics into abstractive text summarization.

The way these stylistic aspects are incorporated into the sequence-to-sequence summarization framework, depends on how these aspects are defined. For example, simplicity of text can be defined at a lexical level based on the word-frequency in a simple corpus as defined in \citebvs{paetzold2015lexenstein}. Krishna et al. \citebvs{krishna2018vocabulary} extend this towards generating simple summary by modifying the decoder probability in Eq.~\ref{eq:pw}. They incorporate simplicity by modifying the beam search decoder to choose contextual replacement with words that are simpler; defining simplicity as, 
\begin{equation}
Simplicity(S) = \frac{1}{m} \sum_{i=1}^{m}\frac{f(s_{i})}{1000},\label{eq:simplicity}
\end{equation}
where $f(s_i)$ is the frequency of the $i^{th}$ word in the SUBTLEX~\cite{brysbaert2009moving} corpus. Krishna et al. \cite{krishna2018vocabulary} then use a word-to-word affinity/ replacement probabilities to achieve the tailoring. 

However, not all aspects can be defined at a lexical level and hence, it is not always straightforward to modify the decoder probability. For example, readability can be quantified via the Flesch reading-ease score~\cite{flesch1979write} given by, 
\begin{equation}206.835 - 1.015\frac{\textrm{total words}}{\textrm{total sentences}} - 84.6\frac{\textrm{total syllables}}{\textrm{total words}}.\end{equation} 
The Flesch reading-ease score quantifies the difficulty in understanding a passage written in English. Higher scores indicate easier to read passages. It posits that the readability is inversely related to the average number of words in a sentence and to the average number of syllables in a word. Using a partial form of this definition, \citebvs{krishna2018vocabulary} propose to use shorter words (lesser syllables) as a surrogate to reading ease and use it to modify decoder probabilities. However, ignoring the first component of the reading ease makes this an incomplete tailoring. 

Accounting for the first component would require the whole sentence to be generated before providing any feedback to the model on its generated readability. We propose to use a reinforcement learning based framework to incorporate such complex objectives requiring the generation of the complete/partial output for feedback generation. %Note that, for this, the model would need an external feedback, in the form of an oracle, which provides feedback to the learning process for tuning towards these constraints based on the partial sequence generated. 
Recently, reinforcement learning frameworks have been successfully used to optimize text generation for content-based metrics, e.g. Rouge scores for summarization in \cite{paulus2018deep}, CIDEr scores for image captioning in \cite{rennie2017self}. We extend these to propose a reinforcement loss for stylistic elements as an additional term along with cross entropy using the Self Critical Sequence Training (SCST)~\cite{rennie2017self} algorithm. 

Given an input article word sequence $x$, and a corresponding ground truth summary $y^{*}=\{y^{*}_{1}, y^{*}_{2}, ..., y^{*}_{T}\}$, the pointer generator framework optimizes the negative log-likelihood objective function is given by, 
\begin{equation}
L_{nll}=-\sum_{t=1}^{T}log\big[ p(y^{*}_{t}|y^{*}_{1},...,y^{*}_{t-1},x)\big].\end{equation} 
For providing explicit feedback on the stylistic characteristics, two output sequences are generated at the time of training: sampled sequence $y^{s}$ and baseline sequence $y^{b}$. 
We generate $y^{s}$ by sampling from the $p(y^{s}_{t}|y^{s}_{1},y^{s}_{2},..,y^{s}_{t-1},x)$ distribution at each time step, and $y^{b}$ by greedily choosing the word with maximum probability from the output distribution at each time step. The SCST algorithm defines a loss term $L_{nll}$ with a reward for the target style characteristics, 
\begin{equation}
L_{rl}=\big[r(y^{b})-r(y^{s})\big]\sum_{t=1}^{T}log\big[ p(y^{s}_{t}|y^{s}_{1},..,y^{s}_{t-1},x)\big],
\end{equation}
where $r(.)$ is the reward function - based on the target style characteristic to be optimized. Optimizing $L_{rl}$ improves the expected reward of the generated output. The final loss is a linear combination of $L_{nll}$ and $L_{rl}$ given by, 
\begin{equation}
L=(1-\alpha) \cdot L_{nll}+(\alpha) \cdot L_{rl},\end{equation} 
where $\alpha$ governs the strength of RL based loss term. Reinforcement learning allows the loss function to include any non-differentiable metric in the form of rewards, which can be leveraged to optimize on our complex stylistic aspects directly. The Self Critical Sequence Training approach also helps in dealing with a exposure bias~\cite{paulus2018deep}, a limitation in teacher forcing algorithm for training recurrent neural networks. By using sampled sequences, the model is exposed to its own distribution, learning to generate in accordance with such global meta properties. 

To evaluate this methodology towards incorporating such characteristics, we use our setup to improve on the readability and simplicity of the generated summaries and incorporate the corresponding metrics into the learning algorithm directly as a reward function using the Reinforcement learning based loss function $L_{rl}$. 
To gain more insights on appropriate methods of tailoring stylistic characteristics, we compare our RL-based approach against the pointer-generator method from~\citebvs{see2017get}, the use of vocabulary tokens adapted from~\citebvs{fan2017controllable} and with modifying word-to-word affinity probabilities adapted from~\citebvs{krishna2018vocabulary}.

To adapt token based approach for readability~\citebvs{fan2017controllable}, we define two tokens: ``Not Readable'' and ``Readable'' based on whether the readability of the ground truth summary was less or more as compared to the median value of $58.58$ in the training dataset. We also evaluate against the lexical level modifications suggested by ~\citebvs{krishna2018vocabulary} and use their \textbf{VoTing} method to modify the generation probabilities by promoting the generation of shorter words over their longer synonyms. Finally, for our RL-based approach, we observe that training using the reinforcement loss is extremely slow, owing to the computation of sampled and greedy sequences along with the teacher forced outputs. Hence, we use the \textbf{PGen} model, pre-trained on \textbf{CNN-DM} dataset as the initialization point, i.e. training with $\alpha=0$. Then, we train for $10,000$ more iterations with a fixed $\alpha=0.9$. 

Similarly, for simplicity, we leverage the work by ~\citebvs{krishna2018vocabulary} to measure simplicity based on Eq. \ref{eq:simplicity} and establish the baselines similar to readability above. Our \textbf{RL-based} method directly uses this simplicity score as the reward function. For the token based approach, we divide the ground-truth summaries into two classes, ``Not Simple'' and ``Simple'' by thresholding at the median, observed to be $8.36$ in the training dataset. 

Similar to content-specific characteristics, we use ROUGE 1, ROUGE 2 and ROUGE L F-scores to evaluate the overlap between the generated and ground-truth summaries. To evaluate whether the models are able to capture our readability definition, we report the average Flesch reading-ease score for the generated summaries. For simplicity, we report the corresponding average simplicity score. Note that our objective is to understand whether the methods are able to capture a given definition for style-specific characteristics. Therefore, we have evaluated the tailoring based on the defined target metrics itself.

Table \ref{tab:readable_simplicity_results} summarizes our experiments for readability and simplicity. We observe a trade-off between the use of simpler/readable words from the vocabulary and the generation quality as captured by ROUGE metric primarily because of the deviation towards more simpler or readable words from the ones in reference summaries. The proposed RL-based approach is better able to capture readability, achieving higher average scores over all other approaches. However, it is not the best model for simplicity, where the lexical modifications at the decoder beats the RL method. This suggests that where the entire sequence needs to be generated to measure the stylistic aspect (like readability) it is useful to resort to RL-based frameworks. However, when the stylistic aspect can be measured lexically, decoder modifications perform better. There exist works in the reinforcement learning literature that have explored Actor-Critic methods~\cite{bahdanau2016actor} to provide intermediate feedback to the model even before generating the complete output sequences via partial rewards. Exploring such techniques to tackle simpler to complex definitions for style-specific constraints is a topic for future work.

Also note that token-based frameworks have a mixed results since it heavily relies on the diversity of training data without any explicit signal - hence might not be suited for stylistic aspects unless the training data contains sufficient diversity. It is possible to train a joint model which can be trained using the feedback on both ground-truth summaries in the data and the sequences sampled from the output distributions in a token based framework - which is a subject of further research.

\begin{table}[bth]
	\centering
	\scriptsize{
	\begin{tabular}{|c|||c|c|c||c|||c|c|c||c|||}
		\hline
		& \multicolumn{4}{c|||}{\textbf{Readability}}& \multicolumn{4}{c|||}{\textbf{Simplicity}} \\\hline
		\textbf{Method} & \multicolumn{3}{c||}{\textbf{ROUGE F-score}} & \textbf{Readability}& \multicolumn{3}{c||}{\textbf{ROUGE F-score}} & \textbf{Simplicity}\\ \hline
		& \textbf{1} & \textbf{2} & \textbf{L} & & \textbf{1} & \textbf{2} & \textbf{L} &\\ \hline
		\textbf{PGen} \cite{see2017get} & $\mathbf{36.98}$ & $\mathbf{15.90}$ & $\mathbf{33.56}$ & $53.17$ & $\mathbf{36.98}$ & $\mathbf{15.90}$ & $\mathbf{33.56}$  & $9.37$\\ \hline
		\textbf{Token-based} \cite{fan2017controllable} & $36.84$ & $15.85$ & $33.56$ & $55.35$ & $35.70$ & $15.17$ & $32.50$  & $9.58$\\ \hline
		\textbf{VoTing} \cite{krishna2018vocabulary} & $36.91$ & $15.72$ & $33.48$ & $54.10$& $33.42$ & $13.03$ & $30.3$  & $\mathbf{12.90}$ \\ \hline
		\textbf{RL-based} & $35.70$ & $15.35$ & $32.71$ & $\mathbf{57.50}$ & $31.95$ & $13.05$ & $29.12$ & $10.70$\\ \hline
	\end{tabular}}
	\caption{\label{tab:readable_simplicity_results} Performance of the proposed approach in improving the readability and simplicity of generated summaries.\vs\vs\vs\vs}
\end{table}

\begin{table}[bth]
	\begin{center}\small
		\begin{tabularx}{\linewidth}{|X|}
			\hline
			\textbf{Article(47.18):}  the killing of an employee at wayne community college in goldsboro , north carolina , may have been a hate crime , authorities said tuesday . investigators are looking into the possibility , said goldsboro police sgt. jeremy sutton . he did not explain what may have made it a hate crime . the victim -- ron lane , whom officials said was a longtime employee and the school 's print shop operator -- was white , as is the suspect . lane 's relatives said he was gay , cnn affiliate wncn reported . the suspect , kenneth morgan stancil iii , worked with lane as part of a work-study program , but was let go from the program in early march due to poor attendance , college president kay albertson said tuesday . on monday , stancil walked into the print shop on the third floor of a campus building , aimed a pistol-grip shotgun and fired once , killing lane , according to sutton . stancil has tattoos on his face ...\\ \hline
			\textbf{Reference(48.5):} relatives of wayne community college shooting victim say he was gay , local media report . the suspect had worked for the victim but was let go , college president says .
			the suspect , kenneth morgan stancil iii , was found sleeping on a florida beach and arrested .\\ \hline
			\textbf{PGen(8.23):} wayne community college , north carolina , may have been a hate crime , authorities say . investigators are looking into the possibility , said goldsboro police sgt. jeremy sutton . investigators are looking into the possibility , said goldsboro police sgt. jeremy sutton .\\ \hline
			\textbf{RL-based(50.12):} the killing of an employee at wayne community college may have been a hate crime . the suspect , kenneth morgan stancil iii , worked with lane as part of a work-study program . he has no previous criminal record , authorities say .\\ \hline
		\end{tabularx}
	\end{center}
	\caption{Sample output summary generated by incorporating readability as a reward function, along with baseline and reference summaries on an instance from \textbf{CNN-DM-mixed} dataset. The numbers in brackets refer to the corresponding readability scores. We show just the top few sentences in the input article, in the interest of space.}
	\label{tab:readability_examples}
\end{table}

\begin{table}[bthp]
	\begin{center}\small
		\begin{tabularx}{\linewidth}{|X|}
			\hline
			\textbf{Article:} hong kong -lrb- cnn -rrb- six people were hurt after an explosion at a controversial chemical plant in china 's southeastern fujian province sparked a huge fire , provincial authorities told state media . the plant , located in zhangzhou city , produces paraxylene -lrb- px -rrb- , a reportedly carcinogenic chemical used in the production of polyester films and fabrics . the blast occurred at an oil storage facility monday night after an oil leak , though local media has not reported any toxic chemical spill ...  \\
			\textbf{Summary:} ...five out of six people were \textbf{hurt}\textit{(injured)} by broken glass and have been sent to the hospital for treatment . \\ \hline
			\textbf{Article:} -lrb- cnn -rrb- debates on climate change can break down fairly fast . there are those who believe that mankind 's activities are changing the planet 's climate , and those who do n't . but a new way to talk about climate change is emerging , which shifts focus from impersonal discussions about greenhouse gas emissions and power plants to a very personal one : your health ...   \\
			\textbf{Summary:} ...it 's easy to brush aside debates involving \textbf{big}\textit{(major)} international corporations , but who would n't stop to think and perhaps do something about their own health\\ \hline
		\end{tabularx}
	\end{center}
	\caption{Sample simplified summaries generated by the proposed approach. Words in bold show the use of simpler summaries generated by our approach, while the words in italics are those picked up by the baseline model.}
	\label{tab:simple_examples}
\end{table}

Table \ref{tab:readability_examples} shows the generated output summaries for the RL-based approach and \textbf{PGen} baseline model, on an instance from \textbf{CNN-DM} dataset, where our RL-based method achieves better readability scores using shorter sentence constructs. Such a framework can be used to teach the model on sentence level characteristics required to achieve a target style. Similarly the summaries generated by VoTing based approach is shown in Table \ref{tab:simple_examples} - the generated summaries uses simpler words in the summaries, such as `big' in place of `major' and `hurt' in place of `injured'. In a similar manner, careful modifications of the generated probabilities while decoding can be used to incorporate various other stylistic aspects defined at a lexical level.

\section{Conclusions}
In this work, we study a variety of constraints which may be imposed while generating abstractive summaries of a given input article by categorizing these constraints as either content-specific, which govern what content needs to be generated and style-specific, which govern various stylistic expressions in these outputs. Our experiments indicate that the content-based characteristics can be tailored in the summary via explicitly or implicitly tuning the attention to focus on relevant parts of the network. Approach to tailor stylistic constraints depends on the nature of definition - characteristics defined at lexical level can be tuned better by modifying decoder probabilities during beam search. More complicated metrics can be tuned by using reinforced rewards in the loss function.

\bibliographystyle{splncs}
\bibliography{pco}
\end{document}